\documentclass{article}
\usepackage{iclr2027_conference,times}

\usepackage{latexsym}

\usepackage[T1]{fontenc}

\usepackage[utf8]{inputenc}

\usepackage{microtype}

\usepackage{graphicx}
\usepackage{hyperref}
\usepackage{url}
\hypersetup{hidelinks}

\usepackage{amsmath,amssymb}
\usepackage{booktabs}
\usepackage{xcolor}
\definecolor{DropRed}{RGB}{192,48,48}
\definecolor{RiseAmber}{RGB}{184,128,0}
\newcommand{\abldrop}[1]{\textcolor{DropRed}{\scriptsize$\downarrow$\,#1}}
\newcommand{\abltokup}[1]{\textcolor{RiseAmber}{\scriptsize$\uparrow$\,#1}}
\newcommand{\tokdown}[1]{\textcolor{green!60!black}{$\downarrow$#1}}
\usepackage{multirow}
\usepackage{tabularx}
\usepackage{array}
\usepackage{placeins}

\AddToHook{env/table/begin}{\setlength{\belowcaptionskip}{\baselineskip}}

\title{Competence-Aware Orchestration for LLM-Based Multi-Agent Systems}

\author{
Songmiao Wang \quad
Yuchen Wang \quad
Chenglin Yu \quad
Hongxia Yang \quad
Ming Li\thanks{Corresponding author: \href{mailto:ming.li@polyu.edu.hk}{ming.li@polyu.edu.hk}}\\
The Hong Kong Polytechnic University
}

\hypersetup{
  pdftitle={Competence-Aware Orchestration for LLM-Based Multi-Agent Systems},
  pdfauthor={Songmiao Wang, Yuchen Wang, Chenglin Yu, Hongxia Yang, Ming Li}
}

\iclrfinalcopy

\begin{document}
\maketitle
\begin{abstract}
LLM-based multi-agent systems (MAS) have emerged as a promising paradigm for enhancing complex reasoning through role specialization and collaboration, yet a key challenge remains: \emph{how can collaboration be adapted to a backbone's competence on the current query?} This problem matters because more collaboration is not always better---richer coordination may help on difficult cases, but it also incurs additional token cost and redundant interactions. We propose a two-stage competence-aware framework for automatic MAS orchestration. It first models the LLM backbone's task-conditioned competence for the current query, and then uses this signal to adapt role activation and the communication graph. Across six benchmarks, our method achieves an average score of 93.81\%. Experiments with six LLM backbones of varying competence levels further show that our method adjusts collaboration according to the inferred competence: for higher-competence backbones, it improves average accuracy from 87.22 to 89.97 while reducing token cost by 36.26\%; for lower-competence backbones, it yields a further average gain of 4.58 points over existing methods. The same trend also holds on unseen backbones.
\end{abstract}

\section{Introduction}
Recent advances in large language models (LLMs) have greatly expanded the use of Multi-Agent Systems (MAS) for reasoning, code generation, and complex problem solving \citep{MetaGPT,Autogen,Chatdev,Agentverse}. Related progress in prompting, search, self-refinement, ensembling, and text-based optimization has further improved LLM reasoning and adaptation \citep{MedPrompt,shinn2023reflexion,Llm-blender,yuksekgonul2024textgrad,liang2024encouraging}. By assigning specialized roles and coordinating their interactions, LLM-based MAS can outperform direct single-agent reasoning \citep{LLM-Debate,MultiPersona-Debate,DyLAN}. However, their effectiveness depends critically on collaboration topology and communication pattern: different tasks, and even different instances, may benefit from different forms and scales of collaboration \citep{G-Designer,Masrouter,Puppeteer}. Moreover, more collaboration is not always better: richer role specialization and denser communication may help on difficult cases, but also introduce additional token cost, latency, and redundant interactions \citep{AgentPrune,AgentDropout}. This motivates a competence-aware perspective on automatic MAS construction: \emph{How can collaboration be adapted to a backbone's competence on the current query?}

\begin{figure}[!t]
    \centering
    \includegraphics[width=\linewidth]{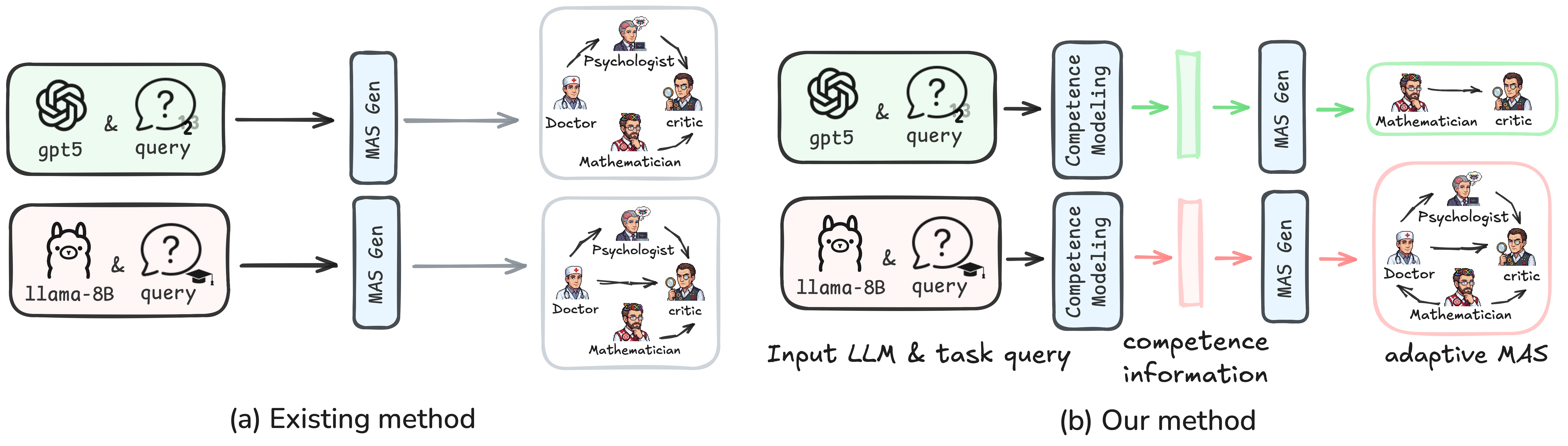}
    \caption{Comparison between existing MAS construction methods and our method.}
    \label{fig:fig1}
\end{figure}

Recent studies have moved from manually designed pipelines and hand-crafted role-based collaboration patterns \citep{MetaGPT,Autogen,Chatdev,LLM-Debate,MultiPersona-Debate} to frameworks that automatically optimize or dynamically generate multi-agent structures \citep{DyLAN,Gptswarm,ADAS,Aflow,G-Designer,MaAs,Masrouter,Puppeteer,MaAS2}. By generating role configurations and communication topologies from task descriptions, query content, candidate roles, or runtime feedback, these methods make MAS increasingly task-aware and adaptive \citep{Aflow,Puppeteer}. Yet the main focus has been on organizing agent collaboration around task semantics, while relatively less attention has been paid to explicitly conditioning its scale and structure on the backbone's competence for the current query. We argue that the utility of collaboration is not fixed: it depends on the backbone's standalone capability, the difficulty of the query, and the coordination overhead introduced by the system. When the backbone's standalone competence is high on a query, extra roles and intensive communication may add cost with little benefit \citep{AgentPrune,AgentDropout}; in contrast, weaker backbones or harder queries may benefit substantially from richer coordination and verification. This pattern also appears in broader agentic benchmarks \citep{MacNet,scaling}. Motivated by these observations, we aim to adapt both \textbf{role participation} and \textbf{communication structure} to the backbone's \textbf{task-conditioned competence}.

To this end, we propose a competence-aware method for automatic MAS construction. Instead of mapping a query directly to a collaboration structure, our method first constructs a task-conditioned competence representation for the given backbone, and then uses it as a control signal for role activation and communication graph construction.

Our contributions can be summarized as follows:
\begin{itemize}
    \item We design a competence modeling mechanism that combines a reusable backbone capability fingerprint with query-specific task information to derive a backbone-aware, query-conditioned competence representation.
    
    \item We develop a competence-aware orchestration method that uses this representation to configure the multi-agent structure at two levels---role activation and communication graph generation---adapting team composition and coordination topology to both the backbone and the current query.
    
    \item Experiments across multiple benchmarks and backbones show that our method improves overall performance, reduces collaboration token usage on strong backbones, and brings larger gains on weaker backbones, with the same trend also observed on unseen backbones.
\end{itemize}

\section{Related Work}

\paragraph{Static multi-agent collaboration and design.}
Early LLM-based multi-agent systems mainly relied on manually designed collaboration templates, including sequential pipelines and debate-style communication \cite{LLM-Debate}. Meanwhile, reasoning can also be improved within a single model through methods such as chain-of-thought \cite{CoT} and self-consistency \cite{SC}. To reduce manual engineering in multi-agent systems, subsequent work began to automate structure design, including workflow optimization methods such as \textsc{AFlow} \cite{Aflow} and \textsc{DSPy} \cite{DSPy}, as well as graph-based approaches such as \textsc{GPTSwarm} \cite{Gptswarm}, \textsc{MacNet} \cite{MacNet}, \textsc{AgentPrune} \cite{AgentPrune}, and \textsc{AgentDropout} \cite{AgentDropout}. These methods improve system design by searching, optimizing, or pruning collaboration structures before deployment. However, whether hand-crafted or automatically optimized, they remain largely input-independent and follow a \emph{learn once, deploy many} paradigm.

\paragraph{Dynamic orchestration.}
More recent studies move beyond static structures and dynamically organize collaboration at inference time. Rather than reusing one fixed design, these methods instantiate or route agent interaction according to the current query or runtime setting. Representative examples include \textsc{G-Designer}, which generates query-specific communication graphs \cite{G-Designer}; \textsc{MasRouter}, which routes topology, roles, and models \cite{Masrouter}; and \textsc{MaAS}, \textsc{Puppeteer}, \textsc{CARD}, and \textsc{MaAS2}, which dynamically orchestrate agent workflows under different settings \cite{MaAs,Puppeteer,CARD,MaAS2}. This line of work makes multi-agent systems more responsive to instance-level variation. However, existing methods mainly adapt collaboration to task semantics or runtime context, with limited explicit modeling of the backbone's query-specific competence. Our work uses task-conditioned competence as a signal for adapting role participation and communication.

\begin{figure*}[!t] 
    \centering
    \includegraphics[width=\textwidth]{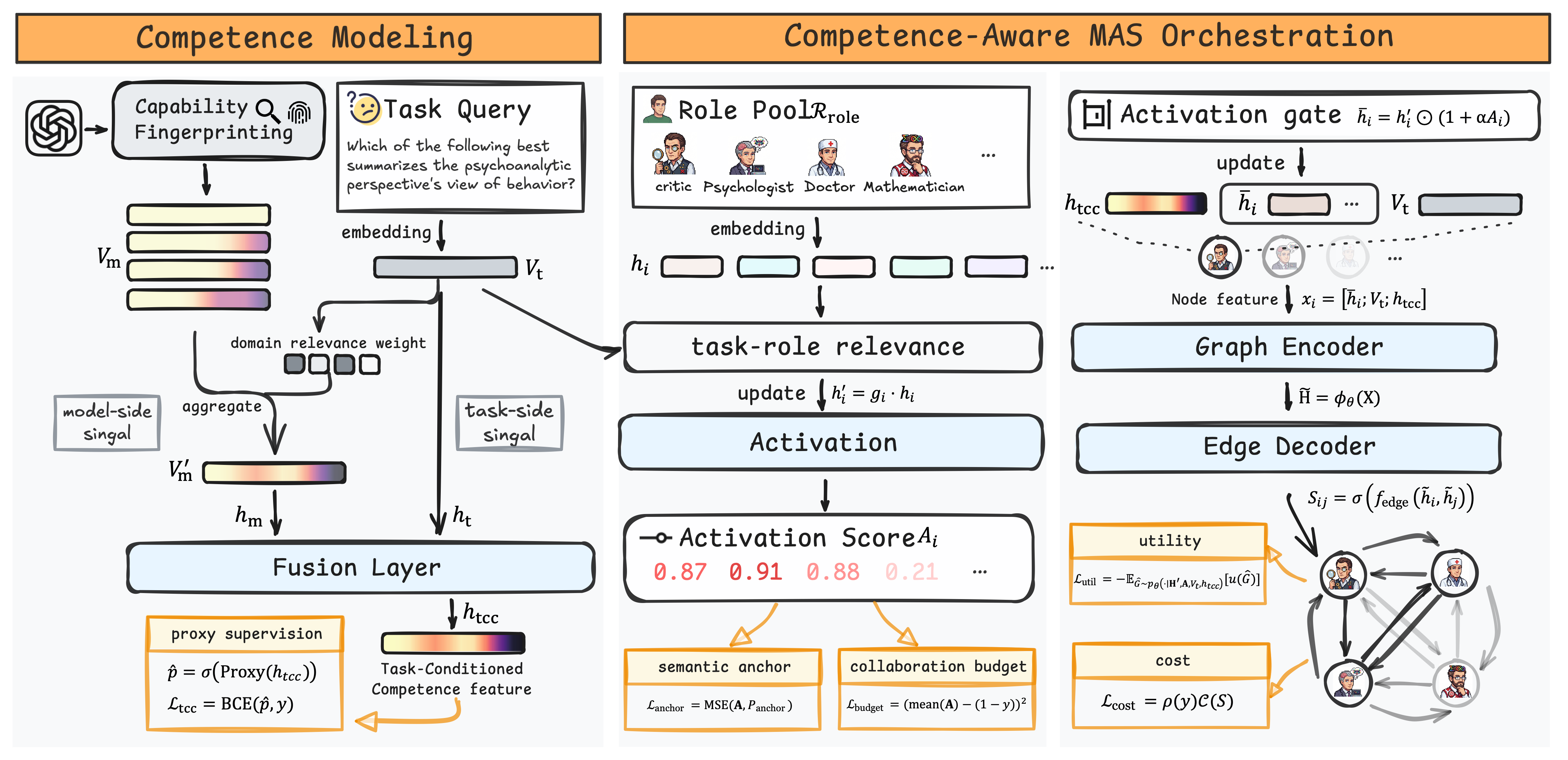}
    \caption{Overview of the proposed competence-aware orchestration framework. Stage I learns a competence representation from the backbone's capability fingerprint and the current query under success-rate supervision. Stage II uses this representation to guide role activation and communication graph generation.}
    \label{fig:main}
\end{figure*}

\section{Method}

\subsection{Overview}

Our method follows a two-stage competence-aware orchestration pipeline. In Stage I, we learn a task-conditioned competence representation from a reusable backbone capability fingerprint and query information. In Stage II, we use this representation to adapt the multi-agent structure at two levels: role activation and communication graph construction.

\subsection{Stage I: Competence Modeling}

\subsubsection{Capability Fingerprinting}

To support downstream query-level competence modeling, we precompute a reusable capability fingerprint for each LLM backbone using a fixed probe set \(\mathcal{P}\). The probes are shared across backbones, organized by capability domain, and ordered within each domain by anchor-calibrated difficulty, yielding a common easy-to-hard scale for direct comparison. Note that this fingerprinting step is performed only once for each backbone: after a backbone \(M\) is first evaluated on \(\mathcal{P}\), its fingerprint \(V_m\) can be reused in all subsequent queries.

We represent the fingerprint as a binary matrix \(V_m\in\{0,1\}^{4\times 12}\), where each row corresponds to one capability domain and the columns within each row are ordered from easy to hard. Let \(\mathcal{D}=\{1,2,3,4\}\) index the domains, and let \(\{x_{d,j}\}_{j=1}^{12}\) denote the fixed probes in domain \(d\). Running \(M\) on each probe \(x_{d,j}\) yields a binary correctness signal \(r(x_{d,j};M)\in\{0,1\}\), from which we construct
\begin{equation}
V_m[d,j] \;=\; r(x_{d,j};M), \quad d\in\mathcal{D}.
\end{equation}
Because probe identities and their within-domain difficulty ordering are shared across backbones, \(V_m\) serves as a directly comparable capability prior for downstream competence modeling: each row summarizes performance in one capability domain, while the column position reflects how far competence extends along the shared difficulty scale. Full probe-set construction details are provided in Appendix~\ref{sec:appendix_probe_set}.

\subsubsection{Task-Conditioned Competence Modeling}
\label{sec:tccm}

To condition the capability prior \(V_\mathrm{m}\) on the current query, we combine it with the task representation to learn a task-conditioned competence representation \(h_{\mathrm{tcc}}\). Query-level success rates provide proxy supervision for learning this representation, which then conditions downstream role activation and communication graph construction.

Given a query \(q\), we first encode it into a task representation \(V_\mathrm{t}\in\mathbb{R}^{d}\). To capture the domain demands implied by \(q\), we introduce a set of learnable domain prototypes \(V_\mathrm{d}\in\mathbb{R}^{4\times d}\), with one prototype for each capability domain in \(V_\mathrm{m}\), initialized from the embeddings of short GPT5-generated domain descriptions. We then compute the similarity between \(V_\mathrm{t}\) and each domain prototype, and normalize the resulting scores with a softmax to obtain domain relevance weights \(w\in\mathbb{R}^{4}\). Using \(w\), we aggregate the domain-wise fingerprint into task-conditioned capability evidence:
\begin{equation}
V_\mathrm{m}'=\sum_{i=1}^{4} w_i \cdot V_\mathrm{m}[i,:].
\end{equation}
Since the columns of \(V_\mathrm{m}\) are ordered by increasing difficulty, \(V_\mathrm{m}'\) provides a query-relevant and difficulty-aware summary of the backbone's capability fingerprint, which is then fused with task semantics to form the final competence representation.

We then project \(V_\mathrm{t}\) and \(V_\mathrm{m}'\) into a shared latent space, denoted by \(h_\mathrm{t}\) and \(h_\mathrm{m}\), and fuse them through an interaction-aware fusion layer:
\begin{equation}
h_{\mathrm{tcc}}=\phi_\mathrm{f}\!\Big(\big[\,h_\mathrm{t};\ h_\mathrm{m};\ (h_\mathrm{t}-h_\mathrm{m});\ (h_\mathrm{t}\odot h_\mathrm{m})\,\big]\Big),
\end{equation}
where \(\odot\) denotes element-wise multiplication, \([\cdot;\cdot]\) denotes concatenation, and \(\phi_\mathrm{f}(\cdot)\) is a learnable fusion network.

We use query-level success rates as proxy supervision for \(h_{\mathrm{tcc}}\). An auxiliary linear head followed by a sigmoid maps this representation to a predicted success rate:
\begin{equation}
\hat{p}=\sigma\big(\mathrm{Proxy}(h_{\mathrm{tcc}})\big).
\end{equation}
The supervision target \(y\in[0,1]\) is the empirical success rate of the corresponding LLM backbone on the query, estimated from repeated trials. Details of the training data construction are provided in Appendix~\ref{sec:appendix_competence_data}. We optimize the model by minimizing the binary cross-entropy (BCE) between the predicted success rate \(\hat{p}\) and the supervision target \(y\).
The prediction head supplies auxiliary supervision for learning competence-related information; downstream role activation and communication graph generation are conditioned on the representation \(h_{\mathrm{tcc}}\).

\subsection{Stage II: Competence-Aware MAS Orchestration}

Given the task-conditioned competence representation produced in Stage I, Stage II adapts the multi-agent structure for the current query through two mechanisms: role activation and communication graph construction. The former assesses how suitable each role is for participation, while the latter determines how the activated roles should communicate. 

\subsubsection{Role Activation}
\label{sec:role_activation}

Given the query representation \(V_\mathrm{t}\) and the task-conditioned competence representation \(h_{\mathrm{tcc}}\), role activation predicts a query-specific activation vector \(A\in(0,1)^N\) over the candidate role pool. It combines task--role relevance with competence-conditioned modulation, and the resulting \(A\) serves as a prior for downstream communication graph construction.

Let the candidate role pool be \(\mathcal{R}_{\mathrm{role}}=\{r_i\}_{i=1}^{N}\), where each role \(r_i\) is associated with a textual description \(d_i\). We encode each description into a role representation \(h_i\in\mathbb{R}^{d_\mathrm{r}}\), and compute a relevance gate based on projected task--role similarity to modulate each role representation:
\begin{equation}
g_i=\sigma\!\left((W_\mathrm{t}V_\mathrm{t})^\top(W_\mathrm{r}h_i)\right),\quad h'_i=g_i\cdot h_i.
\end{equation}
We then predict the final activation score of role \(i\) by conditioning on both the gated role representation and the task-conditioned competence signal:
\begin{equation}
A_i=\sigma\!\Big(\mathrm{MLP}\big([h'_i;\,h_{\mathrm{tcc}}]\big)\Big),\quad i=1,\dots,N,
\end{equation}
yielding \(\mathbf{A}=[A_1,\dots,A_N]\in(0,1)^N\). Here, task--role relevance determines which roles are plausible for the current query, while \(h_{\mathrm{tcc}}\) modulates how strongly they should be activated given the backbone's estimated competence.

Since direct supervision for role activation is unavailable, we train the activator with two weak supervision signals. The first is a semantic anchor derived from task--role similarity, which provides role-level relevance supervision. The second reuses the query-level success-rate supervision from Section~\ref{sec:tccm} to form a soft budget prior that regulates the overall activation scale.

Specifically, we compute the cosine similarity between the projected task representation and each projected role representation:
\begin{equation}
\tilde V_\mathrm{t} = W_\mathrm{t}V_\mathrm{t},\; \tilde h_i = W_\mathrm{r}h_i,\; s_i=\cos(\tilde V_\mathrm{t},\tilde h_i).
\end{equation}
We then apply min--max normalization over the role pool to obtain a query-local semantic anchor target $P_{\text{anchor}}$.
Let \(P_{\text{anchor}}=[P_{\text{anchor},1},\dots,P_{\text{anchor},N}]\in[0,1]^N\). We align the predicted activations \(A\) with this semantic anchor using
\begin{equation}
\mathcal{L}_{\text{anchor}}
=
\operatorname{MSE}\!\big(\mathbf{A}, P_{\text{anchor}}\big).
\end{equation}
This objective provides relative supervision within each query, encouraging higher activation for roles that are more semantically compatible with the task.

To regulate the overall activation scale, we again reuse the success-rate supervision target \(y\in[0,1]\) from Section~\ref{sec:tccm}. We use \(1-y\) as a query-level soft budget prior and regularize the mean role activation toward this target:
\begin{equation}
\mathcal{L}_{\text{budget}}
=
\Big(\operatorname{mean}(\mathbf{A})-(1-y)\Big)^2.
\end{equation}
This term encourages lower mean activation for queries with higher standalone success rates and higher mean activation for those with lower success rates. Together with the semantic anchor, it provides weak supervision for competence-conditioned role selection.

The overall objective is
\begin{equation}
\mathcal{L} = \lambda_{\text{anchor}}\mathcal{L}_{\text{anchor}}+\lambda_{\text{budget}}\mathcal{L}_{\text{budget}}.
\end{equation}
The resulting activation vector \(A\) determines role participation for the current query and serves as input to downstream communication topology generation.

\subsubsection{Graph-based MAS Generation}
\label{sec:mas_topology_prediction}

Building on the competence-aware role activation above, we generate a directed communication graph for the current query. Specifically, given the task representation \(V_\mathrm{t}\), the task-conditioned competence representation \(h_{\mathrm{tcc}}\), the competence-aware role representations \(\mathbf{H}'=\{h_i'\}_{i=1}^{N}\), and the corresponding activation scores \(\mathbf{A}=\{A_i\}_{i=1}^{N}\), we predict a directed communication graph \(G_{\mathrm{com}}=(\mathcal{V}, \mathcal{E}_{\mathrm{com}})\) over the role set \(\mathcal{V}=\{1,\dots,N\}\). Each directed edge represents a potential information-flow path used in multi-round multi-agent interaction. Similar to \cite{G-Designer}, we initialize the graph encoder with an anchor topology, which serves only as a structural prior for message passing.

\paragraph{Task-conditioned graph encoder.}
To predict the query-specific communication topology, we encode each role node using its competence-aware role representation and activation score. Specifically, we use the activation score \(A_i\) to modulate the corresponding role representation:
\begin{equation}
\bar{h}_i = h_i' \odot (1 + \alpha A_i),
\end{equation}
where \(\alpha\) is a scaling coefficient.
We then construct the input feature of each role node by concatenating the modulated role representation with the task representation \(V_\mathrm{t}\) and the task-conditioned competence context \(h_{\mathrm{tcc}}\):
\begin{equation}
x_i = [\,\bar{h}_i;\, V_\mathrm{t};\, h_{\mathrm{tcc}}\,].
\end{equation}
Let \(\mathbf{X}=\{x_i\}_{i=1}^{N}\) denote the resulting node features. A graph encoder \(\phi_{\theta}\) then performs message passing over the initialized role graph and produces contextualized node representations:
\begin{equation}
\tilde{\mathbf{H}} = \phi_{\theta}(\mathbf{X}),
\end{equation}
where \(\tilde{\mathbf{H}}=\{\tilde{h}_i\}_{i=1}^{N}\). In this way, topology prediction is jointly conditioned on role semantics, task information, task-conditioned competence context, and role activation strength.

\paragraph{Edge decoding and graph sampling.}
Based on the encoded node representations, we compute an edge score for each allowable directed edge \((i,j)\in\mathcal{E}_{\mathrm{allow}}\):
\begin{equation}
S_{ij} = \sigma\!\bigl(f_{\mathrm{edge}}(\tilde{h}_i, \tilde{h}_j)\bigr),
\end{equation}
where \(f_{\mathrm{edge}}\) is the edge decoder and \(S_{ij}\in(0,1)\) is interpreted as the activation probability of edge \((i,j)\). The query-specific communication graph \(G_{\mathrm{com}}\) is obtained by selecting edges according to these scores over the allowable edge set. During training, we sample a graph instance \(\hat{G}\) from the predicted edge distribution for rollout and optimization. We optimize the expected utility--cost trade-off. Specifically, the utility term is defined as
\begin{equation}
\mathcal{L}_{\text{util}}
=
-
\mathbb{E}_{\hat{G}\sim p_{\theta}(\cdot\mid \mathbf{H}', \mathbf{A}, V_t, h_{tcc})}
\big[u(\hat{G})\big],
\end{equation}
where \(u(\hat{G})\) denotes the utility obtained from the sampled topology \(\hat{G}\). The communication cost term is defined as
\begin{equation}
\mathcal{L}_{\text{cost}}
=
\rho(y)\,\mathcal{C}(S),
\end{equation}
where \(y\in[0,1]\) is the instance-wise success-rate signal introduced in Section~\ref{sec:tccm}, namely, the empirical success rate of the base model on the current query estimated from repeated trials. We use this signal to modulate the communication penalty and define the communication cost as
\begin{equation}
\rho(y)=\max(y,y_{\epsilon}),\;
\mathcal{C}(S)=\sum_{(i,j)\in\mathcal{E}_{\mathrm{allow}}} S_{ij}.
\end{equation}
Here \(y_{\epsilon}>0\) is a small floor constant, and \(\mathcal{C}(S)\) corresponds to the expected number of active edges under the predicted edge probabilities. This competence-conditioned weight reduces the relative communication penalty on queries with lower standalone success rates, while the utility term rewards task performance. The overall objective is
\begin{equation}
\mathcal{L}_{\mathrm{graph}}(\theta)
=
\mathcal{L}_{\text{util}}+\lambda \mathcal{L}_{\text{cost}},
\end{equation}
where \(\lambda\) balances task utility and communication cost. The utility term is optimized with REINFORCE, while the cost term backpropagates directly through \(S\).

\FloatBarrier
\section{Experiments}
\label{sec:experiments}
\FloatBarrier

\subsection{Experimental Setting}
\paragraph{Datasets and Metrics.}
Following prior work on automatic multi-agent system design and orchestration \citep{G-Designer}, we evaluate our method on six widely used benchmark datasets: MMLU \citep{MMLU}, GSM8K \citep{GSM8K}, MultiArith \citep{MultiArith}, SVAMP \citep{SVAMP}, AQuA \citep{AQuA}, and HumanEval \citep{HumanEval}. We report the standard evaluation metric for each dataset, using accuracy for MMLU, GSM8K, MultiArith, SVAMP, and AQuA, and Pass@1 for HumanEval. Dataset statistics are reported in Table \ref{tab:dataset_stats}.
\paragraph{Baselines.}
Following prior work \citep{G-Designer}, we compare our method with three groups of baselines. The first group consists of \emph{single-agent methods}, including Chain-of-Thought (CoT) \citep{CoT} and Self-Consistency \citep{SC}. The second group includes \emph{multi-agent systems with fixed collaboration patterns}, such as Chain, Tree, Complete Graph, and Random Graph \citep{MacNet}, as well as debate-based interaction instantiated by LLM-Debate \citep{LLM-Debate}. The third group covers \emph{multi-agent methods with dynamic or optimized structures}, including AgentPrune \citep{AgentPrune}, AgentDropout \citep{AgentDropout}, MasRouter \citep{Masrouter}, CARD \citep{CARD}, and G-Designer \citep{G-Designer}.

\paragraph{Implementation Details}
We access GPT models via the OpenAI API, primarily using \texttt{gpt-4o-2024-08-06} (GPT-4o) as the target backbone. Unless otherwise specified, all LLM calls in evaluation use a decoding temperature of 0. For multi-agent methods, we adopt explicit agent profiling with a predefined pool of textual role descriptions, following classical configurations in prior LLM-based MAS work \citep{DyLAN,Gptswarm,yin2023exchange}. We additionally include a summarizer agent to aggregate the dialogue history and produce the final answer. Both task queries and role descriptions are encoded using \texttt{bge-small-en-v1.5} \citep{bge_embedding}, yielding 384-dimensional embeddings. We train Task-Conditioned Competence Modeling, Role Activation, and Graph-based MAS Generation sequentially offline, keeping the other modules frozen at each stage. Deployment reuses the trained model without further parameter updates. For training the graph-generation module, following \citet{G-Designer}, we use 40 queries from each dataset for training. We use dataset-specific training and evaluation splits; their definitions and the construction of capability fingerprints and supervision targets are detailed in Appendix~\ref{sec:appendix}.

\subsection{Performance Comparison}
\label{sec:main_results}

The comparison results are presented in Table~\ref{tab:main}. Under the same backbone setting as all baselines, our method achieves the best average performance of 93.81\% across the six benchmarks, establishing state-of-the-art overall results. In Section~\ref{sec:unseen_models}, we further examine whether this advantage generalizes across different target backbones, including unseen models.

\begin{table}[htbp]
\caption{Performance comparison (\%) on six benchmarks.}
\label{tab:main}
\centering
\resizebox{\textwidth}{!}{
\begin{tabular}{l | c c c c c c | c}
\toprule
\textbf{Method} & \textbf{MMLU} & \textbf{GSM8K} & \textbf{AQuA} & \textbf{MultiArith} & \textbf{SVAMP} & \textbf{HumanEval} & \textbf{Average} \\
\midrule
Vanilla & 80.30 & 82.43 & 71.16 & 92.77 & 86.32 & 71.56 & 80.76 \\
\midrule
CoT & 81.64 {\scriptsize $\uparrow$1.34} & 86.81 {\scriptsize $\uparrow$4.38} & 73.21 {\scriptsize $\uparrow$2.05} & 93.12 {\scriptsize $\uparrow$0.35} & 87.33 {\scriptsize $\uparrow$1.01} & 74.75 {\scriptsize $\uparrow$3.19} & 82.81 {\scriptsize $\uparrow$2.05} \\
SC (CoT) & 83.30 {\scriptsize $\uparrow$3.00} & 81.86 {\scriptsize $\downarrow$0.57} & 75.79 {\scriptsize $\uparrow$4.63} & 94.30 {\scriptsize $\uparrow$1.53} & 88.44 {\scriptsize $\uparrow$2.12} & 79.87 {\scriptsize $\uparrow$8.31} & 83.93 {\scriptsize $\uparrow$3.17} \\
\midrule
Chain & 82.66 {\scriptsize $\uparrow$2.36} & 88.28 {\scriptsize $\uparrow$5.85} & 74.38 {\scriptsize $\uparrow$3.22} & 93.42 {\scriptsize $\uparrow$0.65} & 86.95 {\scriptsize $\uparrow$0.63} & 81.57 {\scriptsize $\uparrow$10.01} & 84.54 {\scriptsize $\uparrow$3.78} \\
Tree & 80.93 {\scriptsize $\uparrow$0.63} & 85.28 {\scriptsize $\uparrow$2.85} & 71.59 {\scriptsize $\uparrow$0.43} & 93.54 {\scriptsize $\uparrow$0.77} & 89.09 {\scriptsize $\uparrow$2.77} & 80.74 {\scriptsize $\uparrow$9.18} & 83.53 {\scriptsize $\uparrow$2.77} \\
Complete & 82.59 {\scriptsize $\uparrow$2.29} & 79.76 {\scriptsize $\downarrow$2.67} & 72.86 {\scriptsize $\uparrow$1.70} & 94.82 {\scriptsize $\uparrow$2.05} & 84.28 {\scriptsize $\downarrow$2.04} & 79.07 {\scriptsize $\uparrow$7.51} & 82.23 {\scriptsize $\uparrow$1.47} \\
Random & 84.12 {\scriptsize $\uparrow$3.82} & 87.09 {\scriptsize $\uparrow$4.66} & 76.89 {\scriptsize $\uparrow$5.73} & 94.14 {\scriptsize $\uparrow$1.37} & 87.56 {\scriptsize $\uparrow$1.24} & 82.67 {\scriptsize $\uparrow$11.11} & 85.41 {\scriptsize $\uparrow$4.65} \\
LLM-Debate & 84.67 {\scriptsize $\uparrow$4.37} & 91.47 {\scriptsize $\uparrow$9.04} & 77.79 {\scriptsize $\uparrow$6.63} & 96.33 {\scriptsize $\uparrow$3.56} & 90.12 {\scriptsize $\uparrow$3.80} & 85.04 {\scriptsize $\uparrow$13.48} & 87.57 {\scriptsize $\uparrow$6.81} \\
\midrule
AgentPrune & 85.40 {\scriptsize $\uparrow$5.10} & 91.32 {\scriptsize $\uparrow$8.89} & 80.52 {\scriptsize $\uparrow$9.36} & 94.91 {\scriptsize $\uparrow$2.14} & 90.17 {\scriptsize $\uparrow$3.85} & 86.94 {\scriptsize $\uparrow$15.38} & 88.21 {\scriptsize $\uparrow$7.45} \\
AgentDropout & 85.71 {\scriptsize $\uparrow$5.41} & 91.47 {\scriptsize $\uparrow$9.04} & 81.12 {\scriptsize $\uparrow$9.96} & 95.18 {\scriptsize $\uparrow$2.41} & 91.41 {\scriptsize $\uparrow$5.09} & 86.24 {\scriptsize $\uparrow$14.68} & 88.52 {\scriptsize $\uparrow$7.76} \\
MasRouter & 89.54 {\scriptsize $\uparrow$9.24} & 94.37 {\scriptsize $\uparrow$11.94} & 87.14 {\scriptsize $\uparrow$15.98} & 92.14 {\scriptsize $\downarrow$0.63} & 85.94 {\scriptsize $\downarrow$0.38} & 90.08 {\scriptsize $\uparrow$18.52} & 89.87 {\scriptsize $\uparrow$9.11} \\
CARD & 86.93 {\scriptsize $\uparrow$6.63} & 94.53 {\scriptsize $\uparrow$12.10} & 89.05 {\scriptsize $\uparrow$17.89} & 89.29 {\scriptsize $\downarrow$3.48} & 86.67 {\scriptsize $\uparrow$0.35} & 87.60 {\scriptsize $\uparrow$16.04} & 89.01 {\scriptsize $\uparrow$8.25} \\
G-designer & 87.14 {\scriptsize $\uparrow$6.84} & 93.76 {\scriptsize $\uparrow$11.33} & 81.61 {\scriptsize $\uparrow$10.45} & 96.61 {\scriptsize $\uparrow$3.84} & 93.18 {\scriptsize $\uparrow$6.86} & 88.32 {\scriptsize $\uparrow$16.76} & 90.10 {\scriptsize $\uparrow$9.34} \\
\midrule
Ours & 92.81 {\scriptsize $\uparrow$12.51} & 95.00 {\scriptsize $\uparrow$12.57} & 87.14 {\scriptsize $\uparrow$15.98} & 99.11 {\scriptsize $\uparrow$6.34} & 96.25 {\scriptsize $\uparrow$9.93} & 92.56 {\scriptsize $\uparrow$21.00} & 93.81 {\scriptsize $\uparrow$13.05} \\
\bottomrule
\end{tabular}
}
\end{table}

\subsection{Ablation Study}
\label{sec:ablation_competence}

Our full method learns its competence representation from two inputs: the task representation $V_\mathrm{t}$ and the capability fingerprint $V_\mathrm{m}$. We evaluate their downstream contribution through three ablated variants. \textbf{w/o task} removes $V_\mathrm{t}$ from the competence module, so its representation is no longer query-conditioned. \textbf{w/o model} removes $V_\mathrm{m}$, so the module no longer has access to a model-specific capability prior. \textbf{w/ desc.} replaces $V_\mathrm{m}$ with free-form model descriptions (e.g., textual summaries of parameter scale and characteristics).

\begin{table}[htbp]
\caption{Downstream ablation of competence modeling.}
\label{tab:ablation_competence}
\centering
\small
\setlength{\tabcolsep}{4pt}
\renewcommand{\arraystretch}{1.08}
\begin{tabular}{lccccc}
\toprule
\textbf{Setting} & \textbf{HumanEval} & \textbf{AQuA} & \textbf{MMLU} & \textbf{Avg.} & \textbf{Tokens / Task} \\
\midrule
Full    & 92.56 & 87.14 & 92.81 & 90.84 & 2360.45 \\
\midrule
w/o task  & 90.91 & 85.71 & 88.24 & 88.29 {\scriptsize \abldrop{2.55}} & 2758.02 {\scriptsize \abltokup{16.84\%}} \\
w/o model & 87.60 & 87.62 & 88.24 & 87.82 {\scriptsize \abldrop{3.02}} & 6010.64 {\scriptsize \abltokup{154.64\%}} \\
w/ desc.  & 88.43 & 85.24 & 89.54 & 87.74 {\scriptsize \abldrop{3.10}} & 4736.68 {\scriptsize \abltokup{100.67\%}} \\
\bottomrule
\end{tabular}
\end{table}

\textbf{Task conditioning improves the performance--cost trade-off.}
Removing the task representation reduces the average score by 2.55 points and increases token cost by 16.84\%. These results support conditioning the competence representation on the current query for downstream orchestration in this setting.

\textbf{The capability prior helps control collaboration cost.}
Removing $V_\mathrm{m}$ causes the largest token increase (+154.64\%) and also lowers the average score by 3.02 points. Under our orchestration design, the backbone-specific capability prior thus supports lower token usage together with higher average performance.

\textbf{Structured fingerprints outperform the evaluated textual descriptions.}
Replacing $V_\mathrm{m}$ with textual descriptions drops the average score by 3.10 points and increases token cost by 100.67\%. This comparison favors the structured fingerprint over these descriptions as a conditioning signal for orchestration.

\subsection{Cross-Backbone Performance and Generalization to Unseen Models}
\label{sec:unseen_models}

\begin{figure}[!b]
    \centering
    \includegraphics[width=0.65\linewidth]{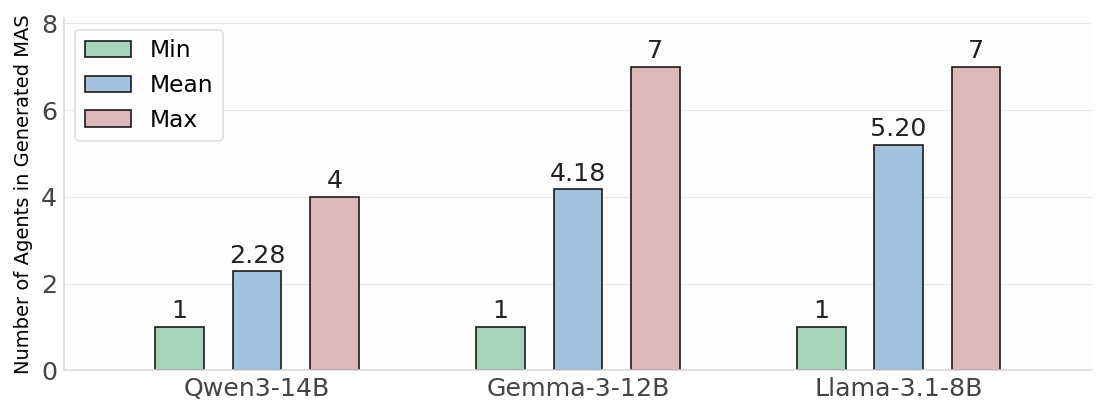}
    \caption{Number of agents in the generated MAS on efficiency-oriented target backbones.}
    \label{fig:activation}
\end{figure}

We evaluate our competence-aware collaboration on MMLU with performance-oriented and efficiency-oriented backbones. We mark as unseen (\(^{\star}\)) backbones used neither as probe-construction anchors nor for competence-supervision data collection. Tables~\ref{tab:unseen_strong} and~\ref{tab:unseen_weak} report accuracy, with Avg. computed across the three backbones in each table. Figure~\ref{fig:activation} shows agent counts across backbones and MMLU queries.

\begin{table}[htbp]
\caption{Results on MMLU with performance-oriented target backbones.}
\label{tab:unseen_strong}
\centering
\small
\setlength{\tabcolsep}{6pt}
\begin{tabular}{lcccc}
\toprule
\textbf{Method} & \textbf{GLM 5.1\(^{\star}\)} & \textbf{Gemini 3.1 Pro\(^{\star}\)} & \textbf{GPT-4o} & \textbf{Avg.} \\
\midrule
Vanilla    & 83.01 & 81.70 & 80.30 & 81.67 \\
G-Designer & 88.89 & 85.62 & 87.14 & 87.22 \\
Ours       & \textbf{89.53} & \textbf{87.58} & \textbf{92.81} & \textbf{89.97} {\scriptsize \tokdown{36.26\%}} \\
\bottomrule
\end{tabular}
\end{table}

\noindent\textbf{On performance-oriented backbones, our method maintains strong performance with substantially lower collaboration cost.}
As shown in Table~\ref{tab:unseen_strong}, our method achieves the best average score on this group, improving over G-Designer from 87.22 to 89.97, while using {\scriptsize \tokdown{36.26\%}} fewer tokens per query after deployment. Capability fingerprinting adds a small one-time cost per backbone, which these savings recover after only about 28 queries on average across this group (Appendix~\ref{sec:appendix_fingerprint_cost}). This result shows that performance-oriented backbones can still benefit from our collaboration strategy, while incurring much lower collaboration overhead.

\begin{table}[htbp]
\caption{Results on MMLU with efficiency-oriented target backbones.}
\label{tab:unseen_weak}
\centering
\small
\setlength{\tabcolsep}{5pt}
\begin{tabular}{lcccc}
\toprule
\textbf{Method} & \textbf{Qwen3-14B} & \textbf{Llama3.1-8B} & \textbf{Gemma-3-12B} & \textbf{Avg.} \\
\midrule
Vanilla 
& 77.79 
& 63.40 
& 66.67 
& 69.29 \\

G-Designer 
& \shortstack[c]{84.31} 
& \shortstack[c]{67.32} 
& \shortstack[c]{70.58} 
& 74.07 \\

Ours       
& \shortstack[c]{\textbf{87.58}} 
& \shortstack[c]{\textbf{70.59}} 
& \shortstack[c]{\textbf{77.78}} 
& \textbf{78.65} \\
\bottomrule
\end{tabular}
\end{table}

\noindent\textbf{On efficiency-oriented backbones, our method brings larger performance gains.}
In Table~\ref{tab:unseen_weak}, our method improves the average score from 74.07 to 78.65, with the largest gain on Gemma-3-12B. Figure~\ref{fig:activation} further shows that the number of agents in the generated MAS varies substantially across target models and tasks.

\noindent\textbf{These improvements also hold on unseen LLM backbones.}
The gains also hold on unseen backbones GLM~5.1\(^{\star}\) and Gemini~3.1~Pro\(^{\star}\). These results support transferring competence-conditioned orchestration beyond the backbones used for probe construction and supervision. We further evaluate the trained model on unseen datasets with an additional unseen backbone in Appendix~\ref{sec:appendix_hard_ood}.

\subsection{Case Study}
\label{sec:case_study}

\begin{figure}[!t]
    \centering
    \includegraphics[width=0.8\linewidth]{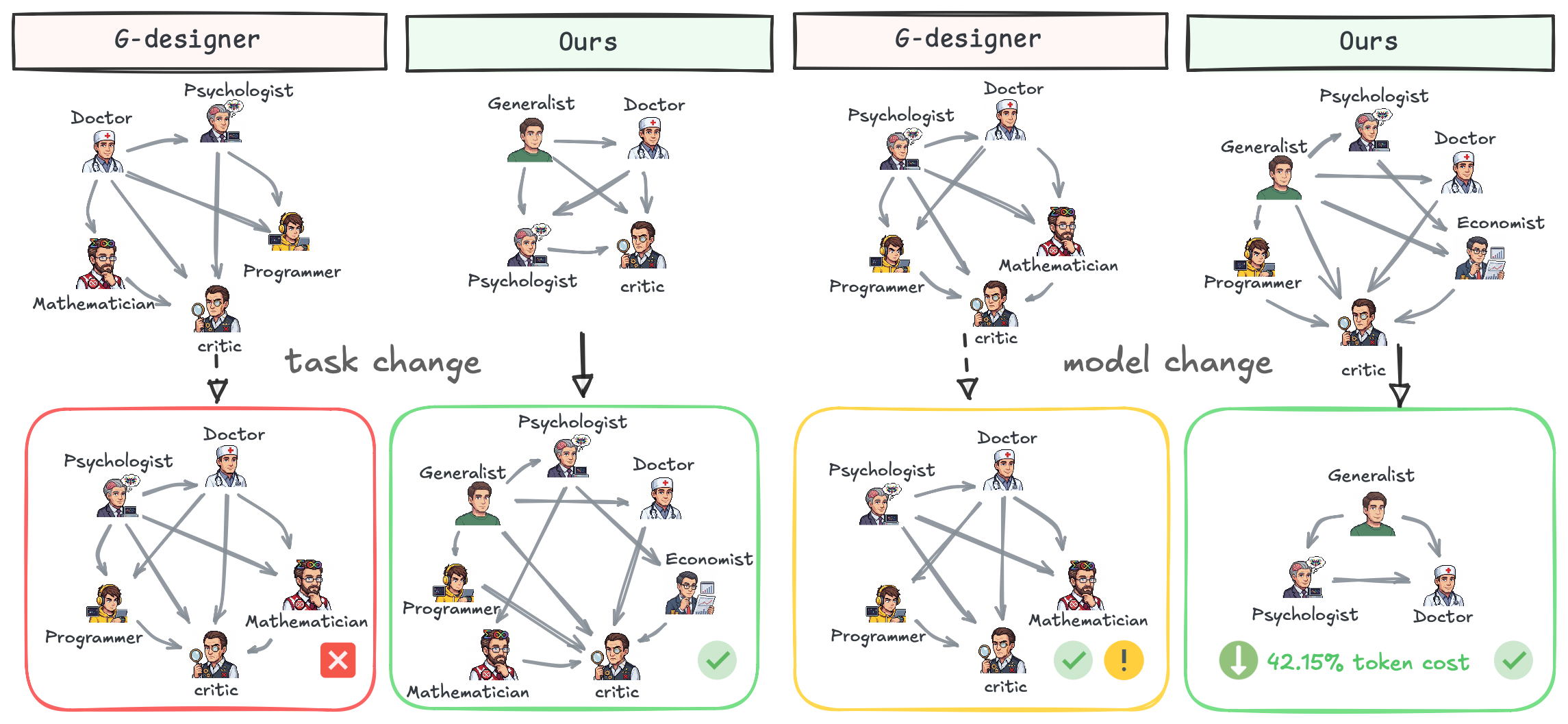}
    \caption{Case studies on MMLU. Left: changing the task while fixing the backbone (Llama-3.1-8B). Right: changing the backbone from Llama-3.1-8B to GPT-4o while fixing the task. check/cross marks indicate answer correctness.}
    \label{fig:case_study}
\end{figure}

We present two MMLU case studies in Figure~\ref{fig:case_study} to show how competence-aware orchestration adapts MAS to different backbone--task settings.

\paragraph{Changing the task while fixing the backbone.}
We fix the backbone to Llama-3.1-8B and compare two psychology questions from \texttt{high\_school\_psychology} and \texttt{professional\_psychology}. The latter requires more specialized expertise. Without explicit competence-aware adaptation, G-Designer makes only limited structural changes and fails on the harder question. In this example, our method activates additional roles and returns a correct answer, illustrating query-dependent adaptation of the collaboration structure.

\paragraph{Changing the backbone while fixing the task.}
We then fix the question and change the backbone from Llama-3.1-8B to GPT-4o. G-Designer continues to instantiate a relatively large collaboration structure. In contrast, our method activates fewer roles with GPT-4o while preserving the correct answer and reducing token cost by 42.2\%. This example illustrates backbone-dependent adaptation with lower collaboration cost.

\FloatBarrier

\section{Conclusion}

In this paper, we study \emph{how to adapt multi-agent collaboration to a backbone's competence on the current query}. We propose a competence-aware framework that learns a task-conditioned representation under success-rate supervision and uses it to guide role activation and communication topology. Experiments across six benchmarks and multiple LLM backbones show that our method improves performance, reduces collaboration cost on stronger backbones, and yields larger gains on weaker ones, with the same trend also observed on unseen backbones.

\clearpage
\subsection*{AI use statement}
Generative AI tools supported this work, including language polishing and code
implementation and debugging. The authors reviewed the AI-assisted materials
and take full responsibility for the final manuscript and results.

\bibliographystyle{iclr2027_conference}
\bibliography{custom}

\appendix

\section{Appendix}
\label{sec:appendix}
\subsection{Probe Set Construction for Capability Fingerprinting}
\label{sec:appendix_probe_set}

\paragraph{Training-query pool.}
Following prior work on automatic MAS design and orchestration \citep{G-Designer,AgentPrune}, which uses \(B' \in \{40,80\}\) training queries per dataset, we adopt the 40-query setting. We randomly sample 40 queries from each of the six benchmark datasets, forming a pool of 240 training-side queries. Probe-set construction and competence-supervision data collection both draw from this pool, without introducing additional distinct task queries. This matches the distinct-query budget, not the offline LLM execution budget.

\paragraph{Dataset-specific splits.}
The dataset loader uses the MMLU development split for training queries and the validation split for evaluation queries. For the other five benchmarks, including HumanEval, fixed dataset-specific index files define the training and evaluation subsets. The training and evaluation index sets in these files are disjoint, and the evaluation loader selects only the designated evaluation indices.

\paragraph{Design goals.}
We construct the probe set to satisfy three requirements. First, it should provide \emph{clear difficulty resolution}, so that backbones of different competence levels can be meaningfully separated rather than collapsing to similar outcomes on the same probe set. Second, it should provide \emph{multi-domain coverage}, so that the resulting capability fingerprint reflects multiple capability dimensions instead of only one narrow skill type. Third, it should remain \emph{compact}, so that one-time fingerprint registration incurs only limited overhead.

\paragraph{Anchor-based difficulty calibration.}
To obtain a stable and interpretable difficulty scale, we evaluate four anchor backbones with progressively stronger standalone capability---\texttt{Llama-3.2-3B-Instruct}, \texttt{Llama-3.1-8B-Instruct}, \texttt{GPT-4o}, and \texttt{GPT-5}---on all 240 sampled queries. For each item \(x\), we record the binary pass vector
\begin{equation}
\mathbf{s}(x) = (s_1, s_2, s_3, s_4), \qquad s_k \in \{0,1\},
\end{equation}
where \(s_k=1\) indicates that the \(k\)-th anchor solves \(x\) correctly, and \(s_k=0\) otherwise.

We order the four anchors from weaker to stronger and retain only items whose pass patterns are monotone under this ordering. Intuitively, if a weaker anchor succeeds while a stronger anchor fails, the item is not suitable for reliable difficulty calibration and is therefore discarded as ambiguous. We additionally remove all-fail items \((0,0,0,0)\), since such items do not help locate the capability range of the anchors and may be confounded by ambiguity, annotation noise, or excessively difficult instances.

The remaining monotone pass patterns define four relative difficulty levels, ranging from easiest to hardest: items solved by all four anchors \((1,1,1,1)\), items solved by all but the weakest anchor \((0,1,1,1)\), items solved only by the two strongest anchors \((0,0,1,1)\), and items solved only by the strongest anchor \((0,0,0,1)\). This yields an empirically grounded difficulty scale shared across candidate backbones.

\paragraph{Domain stratification.}
To ensure that the probe set captures multiple capability dimensions, we further partition the candidate items into four coarse domains according to their source datasets:
\begin{table}[t]
\caption{Domain assignment used in probe-set construction.}
\label{tab:probe_domains}
\centering
\small
\begin{tabular}{ll}
\toprule
\textbf{Domain} & \textbf{Source datasets} \\
\midrule
\texttt{code\_logic} & HumanEval \\
\texttt{quantitative\_reasoning} & AQuA \\
\texttt{basic\_math} & GSM8K, MultiArith, SVAMP \\
\texttt{general\_knowledge} & MMLU \\
\bottomrule
\end{tabular}
\end{table}

\paragraph{Balanced sampling and final probe set.}
Combining the four domains with the four difficulty levels yields 16 domain--difficulty cells. We sample 3 items from each cell, resulting in a final probe set of 48 items. For each domain, the selected probes are ordered from easy to hard as three items from each difficulty level, yielding 12 probes per domain in total. This gives the \(4\times 12\) layout used in capability fingerprinting: the four rows correspond to capability domains, and the 12 columns within each row form a shared easy-to-hard scale.

Overall, this construction provides a compact but structured probe set that simultaneously offers clear difficulty resolution, multi-domain coverage, and low one-time registration cost.

\begin{table}[h]
\caption{Dataset descriptions and statistics.}
\label{tab:dataset_stats}
\centering
\small
\begin{tabular}{l l l r l}
\toprule
Dataset & Answer Type & Metric & \#Test & License \\
\midrule
MMLU & Multi-choice & Acc. & 153 & MIT License \\
GSM8K & Number & Acc. & 1,319 & MIT License \\
MultiArith & Number & Acc. & 600 & Unspecified \\
SVAMP & Number & Acc. & 1,000 & MIT License \\
AQuA & Multi-choice & Acc. & 254 & Apache-2.0 \\
HumanEval & Code & Pass@1 & 164 & MIT License \\
\bottomrule
\end{tabular}
\end{table}

\subsection{Proxy Supervision Data Construction for Task-Conditioned Competence Modeling}
\label{sec:appendix_competence_data}

The goal of task-conditioned competence modeling is to make the fused representation capture how competent a given LLM backbone is expected to be on the current query. Since such query-level competence is not directly annotated, we construct a proxy supervision signal from empirical task performance. The intuition is straightforward: if a backbone succeeds more frequently on a query across repeated trials, it should be regarded as more competent on that query; conversely, a lower success rate indicates lower competence.

We collect this supervision data offline from the same pool of 240 training-side samples used in Appendix~\ref{sec:appendix_probe_set}. After excluding the 48 samples selected as the probe set, we use the remaining 192 samples to build the proxy supervision data. Thus, the probe set is used only for Capability Fingerprint registration, while the non-probe samples are used to estimate query-level competence targets.

For each backbone \(m\) and each non-probe query \(q\), we repeatedly run the backbone \(R=10\) times with decoding temperature set to \(0.2\), and define the supervision target as the empirical success rate:
\begin{equation}
y_{m,q} = \frac{1}{R}\sum_{r=1}^{R} z_{m,q}^{(r)},
\end{equation}
where \(z_{m,q}^{(r)} \in \{0,1\}\) denotes the outcome of the \(r\)-th trial. We use a small non-zero temperature to preserve limited stochastic variation across repeated runs while avoiding excessive randomness in the estimated competence targets. For HumanEval, \(z_{m,q}^{(r)}\) is the pass/fail result under the benchmark unit tests, so \(y_{m,q}\) is an empirical estimate of \texttt{pass@1}. For all other benchmarks, \(z_{m,q}^{(r)}\) is the standard correctness indicator, and \(y_{m,q}\) therefore corresponds to the empirical accuracy on query \(q\).

To ensure that the supervision data covers diverse competence patterns, we construct it using 14 LLM backbones with varying scales, capability levels, and specialization characteristics, including general-purpose, code-specialized, math-specialized, and reasoning-oriented models (Table~\ref{tab:competence_backbones}). GPT-4o serves as a probe-construction anchor (Appendix~\ref{sec:appendix_probe_set}), but is not among these supervision backbones.

For each backbone, we first obtain its Capability Fingerprint from the 48-sample probe set. We then pair this fingerprint with each non-probe query and its corresponding target \(y_{m,q}\), yielding supervised instances of the form
\begin{equation}
\big(\mathrm{Fingerprint}(m),\ q,\ y_{m,q}\big).
\end{equation}
To avoid an overly skewed target distribution and improve the relative coverage of different competence levels, we perform distribution-aware sampling over these candidate instances with respect to \(y_{m,q}\). The resulting training set for task-conditioned competence modeling contains 1060 instances.

These proxy supervision data are used to train task-conditioned competence modeling (\S\ref{sec:tccm}) and provide the supervision basis for downstream role activation (\S\ref{sec:role_activation}). By supervising the fused representation against empirical query-level success outcomes, the model is encouraged to encode competence-relevant information that can directly support role selection and communication graph construction.

\begin{table}[h]
\caption{LLM backbones used for proxy-supervision data construction.}
\label{tab:competence_backbones}
\centering
\small
\setlength{\tabcolsep}{5pt}
\renewcommand{\arraystretch}{1.08}
\begin{tabularx}{\linewidth}{@{}l>{\raggedright\arraybackslash}X@{}}
\toprule
\textbf{Family} & \textbf{Backbones} \\
\midrule
Phi & \texttt{Phi-3.5-mini-instruct} \\
Qwen2.5 & \texttt{Qwen2.5-Coder-7B-Instruct}, \texttt{Qwen2.5-Math-7B-Instruct} \\
DeepSeek & \texttt{deepseek-v3.2} \\
Gemma & \texttt{gemma-3-12b-it} \\
GPT & \texttt{gpt-4o-mini}, \texttt{gpt-5} \\
Llama & \texttt{llama-3.1-8b-instruct}, \texttt{llama-3.2-1b-instruct}, \texttt{llama-3.2-3b-instruct} \\
Qwen3 & \texttt{qwen3-1.7b}, \texttt{qwen3-14b}, \texttt{qwen3-32b} \\
QwQ & \texttt{qwq-32b} \\
\bottomrule
\end{tabularx}
\end{table}

\paragraph{Evaluation of the auxiliary prediction head.}
We evaluate the auxiliary success-rate prediction head used to supervise task-conditioned competence representation learning. We use query-disjoint training and evaluation partitions of the existing success-rate supervision data. All backbone--query pairs sharing the same query are assigned to a single partition, so no evaluation query appears in training or validation. We train the Stage-I module with BCE supervision and select checkpoints using validation MAE. The evaluation covers 153 backbone--query pairs spanning 28 held-out queries from backbones and domains represented during training. Across three training seeds, the predicted success rates achieve a pooled Spearman correlation of \(0.629 \pm 0.022\) with empirical success rates and an MAE of \(0.234 \pm 0.010\) on the \([0,1]\) scale (mean \(\pm\) standard deviation across training seeds). The prediction task provides auxiliary supervision for representation learning, while downstream orchestration is conditioned on the learned representation rather than the scalar success-probability estimate.

\subsection{Amortized Cost of Capability Fingerprinting}
\label{sec:appendix_fingerprint_cost}

\paragraph{One-time registration cost.}
Our efficiency comparison measures deployment-time LLM token usage separately from offline supervision collection and model training. The trained orchestration model is reused without further parameter updates. A new backbone requires a one-time fingerprint registration: evaluating it on the fixed 48-item probe set to obtain \(V_m\), without repeating supervision collection or model training. We measure this registration cost for the three performance-oriented backbones in Section~\ref{sec:unseen_models}: \texttt{Gemini 3.1 Pro}, \texttt{GLM 5.1}, and \texttt{GPT-4o} (Table~\ref{tab:fingerprint_cost}). The amortization below concerns this registration cost, not offline training.

\paragraph{Average cost recovery.}
In Section~\ref{sec:unseen_models}, the average online token usage is 3953.94 tokens/query for G-Designer and 2520.26 tokens/query for our method, yielding an average saving of
\begin{equation}
\Delta_{\text{save}} = 3953.94 - 2520.26 = 1433.68
\end{equation}
tokens per query. Since this saving is averaged over the three performance-oriented backbones, we use it only to compute an average amortization point. Specifically, the average one-time fingerprinting cost across the three backbones is 39,840 tokens, which corresponds to 830.0 tokens per probe over the 48-item probe set. Therefore, the average cost-recovery point is
\begin{equation}
N_{\text{break-even}} = \frac{39840}{1433.68} = 27.79.
\end{equation}
That is, the one-time cost of obtaining \(V_m\) is recovered after only 27.79 queries on average. These results suggest that the probe set remains compact in practice and that the additional registration cost of capability fingerprinting can be amortized quickly in realistic multi-query deployment settings.

\begin{table}[h]
\caption{One-time token cost of Capability Fingerprinting for the three performance-oriented backbones in Section~\ref{sec:unseen_models}.}
\label{tab:fingerprint_cost}
\centering
\small
\setlength{\tabcolsep}{6pt}
\renewcommand{\arraystretch}{1.08}
\begin{tabular}{lcc}
\toprule
\textbf{Backbone} & \textbf{Total Tokens} & \textbf{Avg. Tokens / Probe} \\
\midrule
\texttt{Gemini 3.1 Pro} & 28,576 & 595.3 \\
\texttt{GLM 5.1} & 78,716 & 1640.0 \\
\texttt{GPT-4o} & 12,228 & 254.7 \\
\midrule
Average & 39,840 & 830.0 \\
\bottomrule
\end{tabular}
\end{table}
\subsection{Generalization to Harder Unseen Benchmarks with Gemma4}
\label{sec:appendix_hard_ood}

We further evaluate our trained model on two unseen datasets, GPQA Diamond
\citep{GPQA} and BBEH-mini \citep{BBEH}, with an unseen LLM backbone,
Gemma4. Table~\ref{tab:hard_ood} reports task performance
and collaboration cost.

\begin{center}
\begin{minipage}{\linewidth}
\centering
\parbox{\linewidth}{\refstepcounter{table}\label{tab:hard_ood}%
\centering Table~\thetable: Results on harder unseen
benchmarks with Gemma4. Accuracy is reported in percent.
Green annotations denote token savings relative to G-Designer
for the same benchmark and backbone.}\par\vspace{\baselineskip}
\small
\setlength{\tabcolsep}{3.3pt}
\renewcommand{\arraystretch}{1.03}
\begin{tabular}{llll}
\toprule
\textbf{Benchmark} & \textbf{Backbone} & \textbf{Method} & \textbf{Acc.} \\
\midrule
\multirow{3}{*}{GPQA} & \multirow{3}{*}{Gemma4} & Vanilla & 59.60 \\
 & & G-Designer & 72.73 \\
 & & \textbf{Ours} & \textbf{75.76} {\scriptsize \tokdown{33.6\%}} \\
\midrule
\multirow{3}{*}{BBEH-mini} & \multirow{3}{*}{Gemma4} & Vanilla & 30.22 \\
 & & G-Designer & 45.43 \\
 & & \textbf{Ours} & \textbf{47.61} {\scriptsize \tokdown{33.6\%}} \\
\bottomrule
\end{tabular}
\end{minipage}
\end{center}

\noindent\textbf{Results.}
In the comparisons reported in Table~\ref{tab:hard_ood}, Ours improves
accuracy over G-Designer on Gemma4 by 3.03 points on GPQA and 2.18 points
on BBEH-mini, while reducing tokens by 33.6\% on both benchmarks.
These results support the transfer of competence-conditioned orchestration
to these two unseen benchmarks with Gemma4 without additional training data.

\subsection{Code Repository}
An anonymized code repository is available at
\href{https://anonymous.4open.science/r/Competence-Aware-Orchestration-for-LLM-based-Multi-Agent-Systems-F434}{Competence-Aware Orchestration for LLM-based Multi-Agent Systems}.
\end{document}